\documentclass{article}

\usepackage[final]{neurips_2019}

\usepackage[utf8]{inputenc}
\usepackage[T1]{fontenc}
\usepackage{hyperref}
\usepackage{url}
\usepackage{booktabs}
\usepackage{amsfonts}
\usepackage{nicefrac}
\usepackage{microtype}
\usepackage{graphicx}
\usepackage{xcolor}
\usepackage{lipsum}

\title{
  Information Extraction: An application to the domain of hyper-local financial data on developing countries\\
  \vspace{1em}
  \small{\normalfont Stanford CS224N \{Custom\} Project}  
}

\author{
  Abuzar Royesh \\
  Management Science and Engineering \\
  Stanford University \\
  \texttt{aroyesh@stanford.edu} \\
 \And
   Olamide Oladeji \\
   Management Science and Engineering \\
   Stanford University \\
   \texttt{oladeji@stanford.edu} \\
}

\begin{document}

\maketitle

\begin{abstract}
Despite the need for financial data on company activities in developing countries for development research and economic analysis, such data does not exist. In this project, we develop and evaluate two Natural Language Processing (NLP) based techniques to address this issue. First, we curate a custom dataset specific to the domain of financial text data on developing countries and explore multiple approaches for information extraction. We then explore a text-to-text approach with the transformer-based T5 model with the goal of undertaking simultaneous NER and relation extraction. We find that this model is able to learn the custom  text structure output data corresponding to the entities and their relations, resulting in an accuracy of 92.44\%, a precision of 68.25\% and a recall of 54.20\% from our best T5 model on the combined task.  Secondly, we explore an approach with sequential NER and relation extration. For the NER, we run pre-trained and fine-tuned models using SpaCy, and we develop a custom relation extraction model using SpaCy's Dependency Parser output and some heuristics to determine entity relationships \cite{spacy}. We obtain an accuracy of 84.72\%, a precision of 6.06\% and a recall of 5.57\% on this sequential task. 
\end{abstract}

\section{Key Information to include}
\begin{itemize}
    \item Mentor: Megan Leszczynski
  
\end{itemize}


\section{Introduction}
The use of natural language processing techniques to extract and curate financial data on companies has been on the ascendancy. Companies such as Crunchbase and Pitchbook use Named Entity Recognition, Information extraction, and relation extraction models to track organization activity  such as how much capital companies have raised among other activities from  a variety of online text sources. 
\\
Despite this progress, there have been two major limitations in the advancement of models for this type of application. First, given the fact that many of the models suited to this task are developed for commercial purposes, there is a scarcity of publicly available benchmarks from using open-source models on business text data to track company activity. In addition, a lot of current NER extraction for company activity tracking has focused on text and entities based in developed countries rather than in developing countries. \\
There has been a rising need for data on developing countries obtained through these methods by both development researchers in economics and policy and by corporate entities looking to track business activity in developing markets. Our review of the literature, as described in the next section, highlights two major gaps. First is a gap in literature on annotated financial datasets for developing countries for NER and relation extraction tasks. In addition to this, we find that there is a dearth of research on the performance of state-of-the-art NLP-based NER and information extraction models on financial data generated by non-Western countries. This is important because named entities and financial news writing/reporting linguistic styles may vary significantly as we move from western English-speaking sources to non-Western/developing country sources. 
Our objectives in this project were to curate a data set for this domain, evaluate performance of standard approaches to information extraction on this dataset, and investigate and compare alternative/novel approaches to this task on the data set. In particular, we present herein two approaches to this task:
\begin{itemize}
    \item a novel text-to-text approach to information extraction using T5 - a state-of-the-art transformer model\cite{DBLP:journals/corr/abs-1910-10683}.
    \item Pre-trained CNN-based NER + Shift-Reduce Dependency Parsing + heuristic based extraction: Using a pre-trained (on a large NER model offered SpaCY, a popular industrial NLP library, which uses a CNN architecture. Here, we use spacy's pre-trained English pipeline called: en\_core\_web\_sm which has been trained on written web text (blogs, news, comments) for the NER task and further use. 
\end{itemize}

We achieved promising results with the T5 model which we present, alongside those from our second approach in the results section of this paper.


\section{Related Work}
The task of information extraction, which is to detect extraction relevant information such as relationship between entities from unstructured or structured text data is one that has attracted a lot of research interest in NLP literature. Reviewing the literature, we can identify  four main types of approach to this task:
\begin{itemize}
    \item Unsupervised IE
    \item Semi-supervised IE
    \item Supervised IE
    \item  Rule-based IE
\end{itemize}

As we see in Mykowiecka et al. and Poibeau et al., rule based information extraction focuses on human-designed rules and patterns to detect tuples of tokens/words of interest\cite{mykowiecka2009rule}\cite{poibeau2012multi}. While lacking scalability, this approach has been applied to a lot of domains as seen in literature. For example, in their book chapter, Rivas et al. present an application of rules-based information extraction in the context  of clinical records data\cite{rivas2019information}. In particular, while not providing exact performance descriptions, the authors report that rules were able to successfully  analyze  patients freestyle response to feedback questions. Similarly, Wang et. all present a comprehensive review of information extraction in clinical settings\cite{wang2018clinical}. According to the authors, which surveyed 271 information extraction applications to clinical data, 65\% of them used rules-based information extraction including tasks like extracting information related to diseases like peripheral arterial disease (PAD). There has also been research done to automatically generate information extraction rules as seen by the work of Hanafi et al\cite{hanafi2017seer}. The authors develop a software SEER which uses graphical machine learning  to generate a set of rules to perform information extraction. 
Beyond rules based extraction, the idea of a supervised learning based approach to information extraction has been the most explored approach to this in literature. In particular, as we see in Han and Oleynik (2020), deep learning based approaches have been applied in domains such as medical data to great success\cite{hahn2020medical}. In Nasar, Jaffry, and Malik (2018), a review of the application of supervised learning based approaches to this problem is presented\cite{nasar2018information}. We see that other non-deep learning approaches have also been applied to the domain of extracting information from scientific literature.

One key aspect of information extraction is Named Entity Recognition - NER - which is the automated detection of entities such as names, organizations, or money. NER as an isolated task has received a lot of attention with models obtaining has high as 90\% F-1 score on English-NER. For example, G. Lample et al introduced two neural network architecture driven approaches to NER; one based on LSTM and CRF (Conditional Random Fields), and the other a transition-based approach to construct label segments, achieving with these two what was then a state-of-the-art performance using very limited supervised training data and unlabelled corpus\cite{lample-etal-2016-neural}.CRFs are a type of graphical modeling approach wherein predictions are modeled to better incorporate context as graphical models that implement the dependencies between them. For English NER, using the CoNLL-2002 dataset, their LSTM-CRF model had an impressive F-1 score of 90.94.  Their LSTM-CRF is a key contribution and enabled future applications to domains such as biomedical data as seen in the work of Habibi et. al. (2017)\cite{habibi2017deep} \cite{dernoncourt2017}.
The use of a simple Bidirectional LSTM architecture without CRF has also been applied to great success in NER problems. For example, Chiu and Nichols (2016) achieve an F-1 score of 91.62 on CoNLL-2003\cite{chiu2016named}. 
Beyond supervised learning approaches, unsupervised and semi-supervised based approaoches to NER have also been explored in literature with varying degrees of success. For example, in \cite{10.1007/978-3-030-26072-9_25}, we see that a graphical model, unsupervised learning based approach to NER outperformed supervised learning based CRF models on newswires data.

The use of text-to-text transformer-based models such as T5 and GPT have revolutionalized recent approaches to NLP tasks. These models, which aim to use one model/architecture, via a text-to-text framework to solve a variety of  NLP tasks, from question answering and translation to summarization among other tasks, have gained popularity. T5, an open-source model developed by Google AI is one such text-to-text model \cite{DBLP:journals/corr/abs-1910-10683}.T5 has been applied to custom NER as well as  question answering on the same medical data set\cite{keywood_would_2020}. In our review of the literature, we however did not see an application of a text-to-text model like T5 to achieve simultaneous NER and relation extraction, particularly in the domain of finance. 

The availability of standard data-sets is crucial for the evaluation of the performance of various approaches. While we see that there are several datasets such as the CoNLL-2003, OntoNotes-5 and the i2b2-2014  data-sets which serve as standard datasets for NER training and testing across different languages, and domains like medicine, there are limited publicly available data-sets on finance, and none that we find on finance NER in developing countries \cite{weischedel2013ontonotes} \cite{tjong-kim-sang-de-meulder-2003-introduction}. For finance, Alvarado et al., 2015 provide labelled data for NER on SEC-fillings text\cite{alvarado2015domain}. This dataset is limited and is focused on the United States. It is also the only publicly-available finance related, labelled NER data set we were able to identify during our literature review.

Given the above, it is clear that our review of the literature thus shows clearly a gap which our project sought out to fill with the analysis and data presented in this paper.

\section{Approach}
For our task, for a given sentence, we want the following:
\\
\textbf{Sample Input}: "Apple had a net income of \$9.4 million" 
\newline
\textbf{Sample Output}: (Company:"Apple"), (Monetary\_variable: "a net income"), (Monetary\_value: "\$9.4 million")

We explored two main approaches to this information extraction task: 1) a  text-to-text t5 model for simultaneous entity extraction and relationship tags, and 2) a custom approach which combines outputs from a pre-trained NER model with those from a dependency parser with some heuristics. For the NER and dependency parser in the second approach, we use SpaCy, an open source model which uses residual convolutional neural networks and incremental parsing with what is called bloom embeddings - that is the words are embedded with a Bloom filter such that word hashes are stored in an embedding dictionary as keys for compactness.\cite{spacy}. 
 We describe the sequence of approach thus far in the following paragraphs.

\subsection{Simultaneous Entity Detection and Relation Extraction}
Motivated by the success of large transformer based models such as t5 and GPT-3 in analyzing various NLP tasks through a text-to-text framework, we explored the use of Google's t5 small model, a transformer based model with 60 million parameters, to undertake simultaneous entity detection and relation extraction\cite{DBLP:journals/corr/abs-1910-10683}.  T5 is pre-trained on a massive corpus of online text, which allows the model to be used for several NLP tasks at once. 
We were able to develop scripts to fine-tune (additional training) and predict with the T5 architecture through the HuggingFace's Transformer library and the SimpleTransformers library\cite{DBLP:journals/corr/abs-1910-03771}. 
To get the output entities and relations previously described, we feed in the input text data--custom data we scraped from hundreds of news articles online and manually annotated. We design a custom text output for the text-to-text model. Under this approach, we explore two different text output styles for the T5 model to learn, and also experiment with  different amount of our custom training data. We also vary the T5 model size (from small, to base, to large) and we vary Epochs. We describe our results in the experiments section.

\begin{figure}[hbt]
    \centering
    \includegraphics[width=6cm]{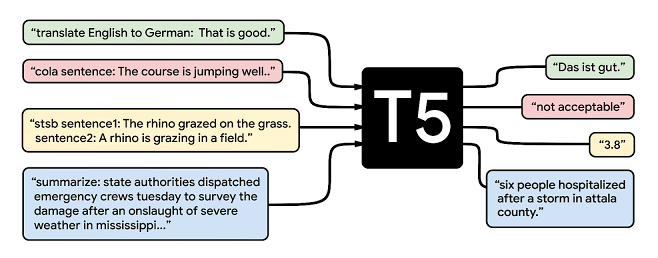}
    \caption{T5 model\cite{DBLP:journals/corr/abs-1910-10683}}
    \label{fig:t5}
\end{figure}

\subsection{Sequential NER + Dependency Parsing and custom heuristics}
Our approach here focused on exploring how we could adapt pre-trained NER and dependency parsing models (on general NER data) to achieve this task. We used SpaCy's model pre-trained on English corpus as loaded via its "en\_core\_web\_md" pipeline, as our base entity detector and dependency parser. We then wrote functions to analyze these outputs using heuristics specially designed for each of our entities of interest. 

Our NER focused on detecting the following five entities: Companies, Persons, Country, Money, and Date.  For these, we needed the overall function to provide an output similar to that generated by the T5 model for easy evaluation. To generate such an output, the relation extraction via dependency parsing and heuristics, focused on establishing the relationship between other entities and the companies detected, if any. For example, for a Money entity detected, we needed to classify it as 'investment' or 'revenue' if it was related to the company. For a Person entity, our objective was to classify this entity as a 'founder' if they were a founder of the company. 

We iteratively experimented with heuristics on detecting the following five pairwise relations:
\begin{enumerate}
    \item company - money
    \item company - date
    \item company - country
    \item company - person.
    \item money - date.
    \item person - country.
\end{enumerate}

Since we needed to classify certain entities, in addition to detecting whether they were related to other entities, we implemented some classification-based heuristics as well. For example, after detecting a money entity and deciding - as per the heuristics - that it was related to another entity such as a company name entity, we needed to classify the money entity detected as either 'revenue' or 'investment' to match the output format recommended by T5. To achieve this, we implemented a function that compared the embeddings of the noun phrase the model predicts as being associated with the money entity to the embeddings of words that are associated with 'investment' or 'revenue'. 

To illustrate this, consider the phrase below: 
\newline
\emph{"Apple had a net income of \$9.4 million"}

The detected money entity is "\$9.4 million" which is successfully related to the detected company named entity "Apple".  To determine whether this money is an investment or revenue, our model also passes "a net income" which is the noun phrase automatically extracted by the heuristic as what relates Apple to the money. Our heuristic classifier function takes in the phrase "a net money" and compares it to the embeddings of a group of revenue-related and investment-related words listed below. 

revenue words = 'revenue','income', 'earnings', 'proceeds', 'returns', 'made'
 \\ 
investment words = 'raised', 'investment', 'received', 'equity'

We then use SpaCy's in-built embedding similarity function to identify what type of word has the maximum similarity to the noun phrase and return revenue or investment if the similarity exceeds 0.5. 

We also explored a similar approach in classifying a detected person entity as a founder or  not. We found this approach worked well on a few examples before evaluating it on the full test dataset.

We also implemented a wrapper function to integrate all the pairwise-relations detected for a paragraph into one output string in the format discussed as with the T5 model.

Overall, to detect the actual pairwise relations listed above, we experimented with a number of heuristics, as informed by our own understanding of the problem. We describe a few in the experiments section.




\section{Experiments}
The following sections provide an in-depth description of our methodology. 

\subsection{Data}
We scraped a prominent African financial news website  (Tech Cabal) to retrieve 400+ news articles from 2019 and 2020 \cite{TechCabal}. We then conducted other pre-processing steps on the dataset to change it to a format suitable for our models. We decided to structure our input at the paragraph-level instead of the sentence-level in order to capture relationships between entities that spanned several sentences. 
In order to use the T5 model, we manually labeled over 6,000 paragraphs in the following format: 
\begin{center}
    \emph{{company name, variable name, variable value, variable date}}\\
\end{center}
where variable name could be one of founder, country, revenue, customers/users, or investment. 

In the cases where there were more than one piece of information in the text, we concatenated them all using a separator in the middle. The following shows one sample input and its corresponding target text: 
\begin{center}
    Input text: f\
    target text: \emph{{Jumia, revenue, €41 million, Q4 2020| Jumia, revenue, €33.7 million, Q3 2020|}}\\    
\end{center}

\subsection{Evaluation method}

To construct a test set, we chose 20\% of the training samples. Since it was possible that two different paragraphs would contain the same information, we made sure that the examples in the test set did not appear in the training set to avoid evaluating the model on a piece of information it had merely memorized. 

In order to properly assess the performance of our models, we used a stringent scoring criteria where we compared each word in the target output to the word generated in the same position in the predicted output. We used exact as well fuzzy similarity scoring (above 90\%) to compare each word. 

\begin{figure}[htp]
    \centering
    \includegraphics[width=6cm]{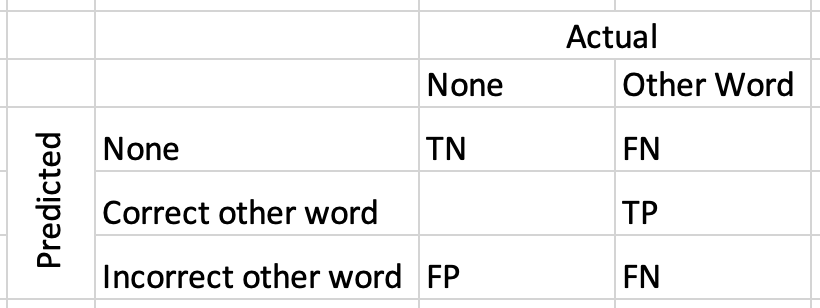}
    \caption{Outcome labels}
    \label{fig:galaxy}
\end{figure}

Using these definitions of true positive, true negative, false positive, and false negative, we calculate accuracy, recall, precision, specificity, and F-score.

\subsection{Experimental details}

\subsubsection{Experimenting with T5}
For the t5-based model, we ran experiments on three levels:
\begin{itemize}
    \item 1) to understand whether we should use the entire training set, use only the training samples that contained information, or both,
    \item 2) to assess which model size had the best performance, and 
    \item  3) to tune the hyper-parameters
\end{itemize}. 
First in order to choose the best model size, we trained each of the small, base, and large models for 1 epoch each to compare performance. Even though t5 has 2 more models that are even larger (3Bn and 11Bn parameters, respectively), we were unable to experiment with them due to the limitations of our virtual machine. 

For the first experiment, we realized that the model had the best performance when we trained it first on the entire training set, and then on a subset of the data containing all the samples that contained information and an equal number of samples that had no relevant information. 

Given that the t5 large model had the best performance over 1 epoch, we then tuned the remaining hyper-parameters on that model. We experimented with decreasing the batch size and increasing the number of epochs to understand how performance would change. For all our experiments we kept the learning rate constant at 1e-4. In order to decide whether whether the model was over or under-fitting the data, we looked at the performance on the training set and compared it to the performance on the test set. We realized that even when we trained the t5 large model for 10 epochs, which took upwards of 5 hours, the training and test accuracy both hovered around 93\%. In an attempt to improve both test and train performance, we decided to run the model for 20 epochs with early stopping to see if performance would improve. However, that led to overfitting on the training data and poor test performance.  

\subsubsection{Experimenting with SpaCY's CNN-based NER and Dependency Parsing Heuristics}

For the sequential NER plus relation extraction, our experiments here focused on implementing and varying several heuristics which we apply to the output of the dependency parser and NER from Spacy. In particular,  we explored various rules for the relation (by navigating the dependency parser tree and factoring the token entities detected) and evaluated their performance on a small subset of the data.

Our experimental process for the heuristic implementation was to work with a small subset of data of about 100 paragraphs, carefully chosen to see different grammatical variations that capture all these pair-wise relations. We then wrote functions with heuristics to detect the entities based on what we see in the subset data. After the implementation of a particular set of heuristics, we re-run on the subset of data and pay attention to the examples on which the heuristic performed poorly. We then iteratively improved on the heuristics with new or modified rules to better address the failed example. After reaching a satisfactory benchmark on the 100 paragraphs, for which we were successfully capturing about 50\% of the relations across all pair-wise mappings, we then ran the model on all the test data to benchmark with the t5.

We describe below, two of our best performing heuristics.
For our Spacy-based model, our best performing  'money' to 'company' and 'money to organization' relation heuristic, were as follows:
\newline
Money-Company mapping: 
 \begin{itemize}
     \item Given a  detected "MONEY" token and its noun chunk, we check if its a an attribute/direct object. If so, we check its left ancestors for a subject and classify that as its related company if its an "ORGANIZATION" entity.
     \item If there is no subject, we also check its verb's children for 'ORGS'.
     \item If it is a prepositional object, we check the head of the preposition and assign this to monetary variable. 
     \item We also find the ancestor verb of this preposition head and check for its 'ORGS' children to identify the company to which it relates.
 \end{itemize}  
 
Company-Date mapping: 
\begin{itemize}
     \item Given a  detected company or "ORGS"/"ORGANIZATION" token and its noun chunk, we check for  all the prepositions to the left subtree and right subtree of the company entity. For all prepositions detected, we check their children tokens to see if we have a "DATE" entity. Any date entity detected is assigned as related to the company/"ORGS" entity.
     \item if the company/"ORGS" entity detected is a direct object, we check for its parent verb and then look at the children of this verb for any date entities. We assign any detected "DATE" entities as related to this company. 
     \item if we have a prepositional object as company/"ORGS" entity with a preposition, we check the preposition head. If the head is a proper noun,  we look at the descendants of the preposition head for "DATE" tags and relate them to the company. We also check the preposition's ancestral verb, and check for "DATE" among the descendants of its ancestral verb. The dates here are mapped to the original company/"ORGS" detected.
 \end{itemize}  

Our best performing heuristic for other pair-wise relations can be found implemented in the code we attach. 
\subsection{Results}
 

The following table summarizes the performance of our select models on the test set. We are only reporting the results from the evaluation based on exact matching.  

\begin{center}
 \begin{tabular}{|c c c c c c c c c|} 
 \hline
 Number & Approach & Model size & Epochs &  Accuracy & Recall & Precision & Specificity & F1-score\\ [0.5ex] 
 \hline\hline
  1 & Heuristic-Baseline & N/A & N/A & 84.72\% & 5.57\% & 6.06\% & 92.03\% & 0.058\\ 
 \hline
 2 & Custom & Large & 1 & 49.58\% & 55.99\% & 9.2\% & 48.98\% & 0.158 \\
 \hline
 3 & Custom & Base & 10 & 92.07\% & 67.41\% & 52.38\% & 94.34\% & 0.590\\
 \hline
 4 & Custom & Large & 10 & 92.44\% & 68.25\% & 54.20\% & 94.68\% & 0.604\\
 \hline
 5 & Custom & Large & 20 & 85.79\% & 32.09\% & 28.03\% & 91.40\% & 0.299\\ [0.5ex] 
 \hline
\end{tabular}
\end{center}

Our best model was the custom transformers-based model run for 10 epochs.

\section{Analysis}
As seen above, our custom-model had an impressive performance on the test set data, despite having relatively small training data and despite us imposing an EXACT matching criteria for the output. We are optimistic that as we get more training data, we will be able to achieve even significantly better model performance. 
To understand our model performance, we took a look at the predictions generated by the baseline heuristic and our best custom model. In doing so, we discovered that our custom model actually performs better than the evaluation results suggest as our current evaluation approach is perhaps very stringent and penalizes the model's results if they are not output in the right order, or are not exact.

Our evaluation method may also have not particularly suited to the baseline heuristic model. For instance, in some cases, the model would pick up names of company affiliates that were not necessarily the founder. In those cases, our evaluation method unfairly would penalize the model for picking up relevant pieces of information. Similarly, our heuristic method would often pick up extra or fewer words, which would be penalized by our stringent scoring criteria.  

In evaluating the predictions of the two models (heuristic-baseline and custom), we also realized that there were some issues with our data entry. For instance, in some cases, the models would pick up true relationships that were mislabeled or not labeled in the training set. We believe that both of our models would benefit greatly from training data of higher quality. 

Overall, we believe that the two approaches provide a promising approach to information extraction in the context of this paper. In particular, we are now highly confident of the transformers model's performance and will be incorporating it into our full pipeline.

\section{Conclusion and Future Work}
In this project, we have highlighted a gap in the use of NLP information extraction techniques on financial data originating from developing countries.
We have curated a labelled dataset of entities and relations from over 6000 paragraphs of financial news sources from developing country sources. We have also presented two different approaches to this task of information extraction in this domain: one a staggered pre-trained CNN-based NER with dependency parsing and heuristics approach, and the other, a novel, text-to-text approach to achieving this task using T5 - a state-of-the-art transformer-based model by Google AI. We find that with the T5 model, despite only training on 6000 paragraphs, we are able to achieve an F-1 score of 0.604 despite using stringent exact matching on a custom string output format. 
\\ We obtained less impressive results with our best sequential NER-Dependency parsing + heuristic model when run on the full data, however, our analysis of why it scored poorly has yielded interesting directions which we will explore. We will also continue collecting more training examples in order to further fine-tune our T5 model-based approach, while also seeing if there are other transformer based approaches (e.g. BERT) that perform well on this task. 
Finally, beyond this class, we will also explore a supervised relation extraction based approach in which a  classifier model will be trained to classify the relationship between two named entities as accurate or inaccurate. 


\bibliographystyle{plainnat}
\bibliography{references}

\appendix


\end{document}